%% file: acl_latex.tex
\documentclass[11pt]{article}

\usepackage[preprint]{acl}

\usepackage{times}
\usepackage{latexsym}
\usepackage[T1]{fontenc}
\usepackage[utf8]{inputenc}
\usepackage{microtype}
\usepackage{inconsolata}
\usepackage{graphicx}
\usepackage{booktabs}
\usepackage{amsmath,amssymb,amsfonts,amsthm}
\usepackage{bm}
\usepackage{multirow}
\usepackage{xcolor}
\usepackage{array}
\usepackage{hyperref}
\usepackage{enumitem}
\usepackage[ruled,vlined]{algorithm2e}
\SetKwInput{KwInput}{Input}
\SetKwInput{KwOutput}{Output}

\newtheorem{proposition}{Proposition}

\newcommand{\method}{\textsc{Fade}}
\newcommand{\E}{\mathbb{E}}
\newcommand{\R}{\mathbb{R}}

\title{\method{}: Factored Ability--Difficulty Filtering for Online RLVR Scheduling}

\author{
  \textbf{Tomas Duncun\textsuperscript{1}},
  Jonah Ellison\textsuperscript{1},
  Mateo Ruiz\textsuperscript{1},
  Hanae Fujita\textsuperscript{2},
  Ingrid Solberg\textsuperscript{2}
  \\
  \\
  \textsuperscript{1}University of Michigan \\
  \textsuperscript{2}University of Edinburgh
}

\begin{document}
\maketitle

\begin{abstract}
Group-based reinforcement learning with verifiable rewards (RLVR) learns only from prompts whose sampled completions disagree: all-pass and all-fail groups produce zero relative advantage and burn generation budget without moving the policy. Adaptive schedulers try to steer rollouts toward informative prompts, yet they rely on per-prompt pass-rate trackers that entangle two quantities evolving at very different speeds---the policy's shifting competence across reasoning skills and each prompt's intrinsic difficulty. When a single exponential average or Beta counter tracks both at once, short memory scrambles the relative ordering of prompts whereas long memory reports stale pass rates. We introduce \method{} (\emph{Factored Ability--Difficulty Estimation}), a bilinear state-space filter that separates the two timescales. \method{} represents each prompt's success logit as the inner product of a whitened $K$-dimensional skill loading with a fast-drifting policy ability vector, minus a slow-shrinking scalar item intercept. Using P\'olya--Gamma augmentation, each minibatch of binary rollout outcomes updates the $K$-dimensional ability posterior in closed form via a rank-$K$ Woodbury step, instantly transferring competence shifts to all unobserved prompts that share those skills while leaving item intercepts stable. From the resulting posterior mean and variance we derive a closed-form utility for the probability of drawing a non-degenerate rollout group. Across five reasoning benchmarks, two base models, and four scheduling paradigms under a matched rollout budget, \method{} consistently improves downstream accuracy and calibration while adding less than $0.6$\,ms per step.
\end{abstract}

\section{Introduction}
\label{sec:intro}

Reinforcement learning with verifiable rewards (RLVR) turns a rule-based answer checker into a policy improvement loop for mathematical and scientific reasoning~\citep{shao2024deepseekmath,guo2025deepseekr1,kimi2025k15}. Group-relative algorithms such as GRPO sample $G$ completions per prompt and standardize the binary rewards within each group~\citep{shao2024deepseekmath,yu2026dapo,zheng2025gspo}. The bottleneck of this recipe is rollout generation, and its central pathology is \emph{group degeneracy}: whenever a prompt is trivial for the current policy ($r_i{=}1$) or far beyond its reach ($r_i{=}0$), all $G$ completions receive the same reward, the group advantage vanishes, and the completions do no gradient work~\citep{yu2026dapo,yao2026gvm,mao2026rlvr}.

\begin{figure*}[t]
    \centering
    \includegraphics[width=\textwidth]{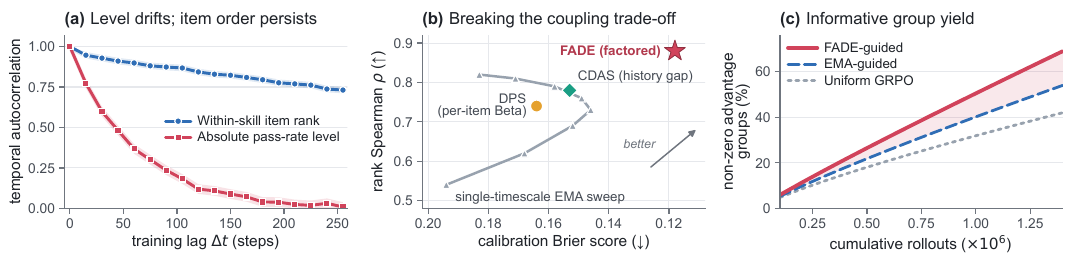}
    \caption{\textbf{Why per-prompt trackers face a coupling trade-off.} \textbf{(a)} Along a GRPO run on NuminaMath, the absolute pass-rate level decorrelates within ${\sim}100$ steps as the policy improves, whereas the relative difficulty ordering of prompts within a skill domain remains correlated over hundreds of steps. \textbf{(b)} Sweeping the decay rate of a per-prompt EMA traces a frontier between calibration error (Brier score) and rank correlation (Spearman $\rho$); per-item Bayesian trackers (DPS, CDAS) sit near that curve, whereas \method{} decouples the two timescales. \textbf{(c)} Supplying factored estimates to a rollout scheduler sustains a higher fraction of non-zero advantage groups across the training budget.}
    \label{fig:motivation}
\end{figure*}

Adaptive RLVR schedulers attack group degeneracy from two angles. Curriculum selectors filter or prioritize prompts whose pass rate sits near the learning frontier~\citep{mao2026dynamics,gao2026pcl,zheng2026greso,kong2026cdas,mao2026rlvr,bae2026online,do2026space}, while budget allocators vary the group size $n_{i,t}$ per prompt to reduce gradient variance or maximize progress~\citep{yao2026gvm,nguyen2026vip,zeng2026cures,zou2026trace,fang2026allocate}. Both families need an online estimate of each prompt's current pass probability $p_{i,t}$ before rollouts are drawn.

Existing trackers treat $p_{i,t}$ as an unstructured per-prompt scalar and run into a fundamental dilemma. On-policy probing before each step is accurate but consumes tens of GPU-hours in extra generations~\citep{yao2026gvm,zeng2026cures,xu2025spo}, while small generalizable prompt predictive models~\citep{qu2026small,gao2026pcl} require auxiliary model training and periodic recalibration. Zero-probe trackers instead smooth past training outcomes for each prompt via exponential moving averages, sliding windows, or dynamics-predictive Beta posteriors~\citep{mao2026dynamics,kong2026cdas,zheng2026greso,nguyen2026vip}. Because a corpus contains $10^5$ prompts and a batch touches only $256$, any individual prompt is revisited once every several hundred steps. Between visits, two things happen at very different speeds (Figure~\ref{fig:motivation}(a)): the policy's overall competence on a mathematical skill drifts quickly, while the fine-grained hardness gap between two problems requiring that skill barely moves. A single per-prompt memory parameter cannot serve both. Fast forgetting tracks recent pass rates on visited items but destroys the relative ranking of rarely visited items; slow forgetting preserves item ordering but reports badly lagging pass probabilities (Figure~\ref{fig:motivation}(b)).

We resolve this tension by factoring the logit of $p_{i,t}$ into a low-rank dynamic policy state and a scalar static item intercept. Specifically, we project frozen prompt embeddings onto $K$ orthonormal whitened skill coordinates $\mathbf{z}_i\in\R^K$ ($K{=}32$) and model the success logit at step $t$ as $\mu_{i,t} = \mathbf{z}_i^\top \boldsymbol{\theta}_t - b_i$. Here $\boldsymbol{\theta}_t\in\R^K$ is a fast-evolving skill proficiency vector shared across the entire corpus, and $b_i\in\R$ is a slow-shrinking intrinsic difficulty offset specific to prompt $i$. Every minibatch of $256$ prompts provides thousands of rollout outcomes that jointly pin down $\boldsymbol{\theta}_t$; updating $\boldsymbol{\theta}_t$ immediately shifts the predicted pass rate of all $N$ prompts in proportion to their skill loadings $\mathbf{z}_j$, even if prompt $j$ has never been sampled. Meanwhile, $b_i$ accumulates evidence over the full run with minimal forgetting, keeping within-skill prompt rankings sharp.

To make inference exact and fast without inner optimization loops, \method{} uses the P\'olya--Gamma augmentation for binomial likelihoods~\citep{polson2013bayesian}. Conditioning on the local curvature parameter turns each step into a conjugate Gaussian Kalman update whose $K\times K$ covariance is maintained in $\mathcal{O}(|B_t|K^2)$ time. Finally, rather than plugging point estimates into heuristic schedulers, we integrate the posterior uncertainty over the binomial group outcome to obtain a closed-form \emph{non-degenerate group utility} $\mathcal{U}_{i,t}$ that directly scores the expected probability of drawing a mixed-reward group.

Our contributions are three-fold:
\begin{itemize}[leftmargin=*,itemsep=2pt,topsep=2pt]
\item We show that online difficulty estimation in RLVR is governed by two separated timescales---fast low-rank skill drift and slow item-level ordering---which single-timescale per-prompt trackers cannot simultaneously capture (\S\ref{sec:intro}, \S\ref{sec:method-model}).
\item We propose \method{}, a P\'olya--Gamma bilinear state-space filter that updates a $K$-dimensional skill state and per-item intercepts in closed form, paired with an uncertainty-calibrated non-degenerate group utility (\S\ref{sec:method}).
\item We prove that \method{}'s Woodbury step is the exact conjugate mode under local P\'olya--Gamma conditioning, that skill sharing strictly contracts predictive variance on unvisited prompts, and that factored filtering dominates single-memory trackers whenever skill drift exceeds item drift (\S\ref{sec:theory}). Across five math benchmarks and four schedulers, \method{} consistently improves RLVR efficiency (\S\ref{sec:experiments}).
\end{itemize}

\section{Related Work}
\label{sec:related}

\paragraph{Curriculum selection and rollout allocation in RLVR.}
Classical curriculum learning orders examples by difficulty or learnability~\citep{bengio2009curriculum,mindermann2022prioritized}. In group-relative RLVR~\citep{shao2024deepseekmath,yu2026dapo,zheng2025gspo}, prompts whose completions all succeed or all fail yield zero advantage, motivating group-prioritized optimization that eliminates ineffective samples~\citep{mao2026rlvr,yu2026dapo}. Recent selectors therefore filter or reweight prompts using dynamics-predictive sampling~\citep{mao2026dynamics}, lightweight prompt predictive models~\citep{qu2026small,gao2026pcl}, competence gaps, or reward dynamics~\citep{zheng2026greso,kong2026cdas,bae2026online,do2026space,sun2025improving,zhang2025grpolead,li2025know,lai2026tacler,jia2026cvpo,pant2026closedloop}. A parallel line allocates non-uniform rollout budgets $n_{i,t}$ across a batch to minimize policy-gradient variance~\citep{yao2026gvm,nguyen2026vip,zeng2026cures}, unify budget allocation across reasoning and agentic RL~\citep{zou2026trace}, or maximize sequential utility~\citep{xiong2025reinforceada,fang2026allocate,jiang2026progressive,chen2026sera,yao2026rollverify,deng2026grpodropout,jiang2026trainingfree,kalarde2026bayes}. Complementary work prunes partial trajectories or rescales advantages~\citep{xu2026prune,xia2026balancing,wang2026whenall,yang2026guidedrlvr,dao2026reapo,son2026rollout,gholipour2026rollout}. All of these schedulers consume an online pass-rate estimate; \method{} replaces that estimator without altering the downstream scheduling rule.

\paragraph{Dynamic difficulty tracking and Item Response Theory.}
Static item response theory (IRT) separates model ability from item difficulty in offline NLP evaluation~\citep{lalor2016building,rodriguez2021evaluation}. During RLVR, however, the policy moves at every step~\citep{zhou2026dare,zhang2026d2evo}. Probing methods estimate current pass rates via extra on-policy rollouts~\citep{yao2026gvm,zeng2026cures,xu2025spo}, trading compute for freshness. Amortized predictive models estimate prompt utility or difficulty from text features~\citep{qu2026small,gao2026pcl,wang2026heapa,zhou2026dare}, offering strong zero-shot generalization when paired with online adaptation. History-based trackers update per-prompt statistics from training rollouts---via dynamics-predictive Beta-Binomial filtering in DPS~\citep{mao2026dynamics}, historical discrepancy~\citep{kong2026cdas}, reward regularity~\citep{zheng2026greso}, or local regression~\citep{nguyen2026vip}---while unified rollout trackers~\citep{zou2026trace}, offline tree structures~\citep{wang2026trees}, bandit curricula~\citep{xu2026predictive}, and lag-corrected replay~\citep{mao2026rlvr,zhang2026reusing,yang2026batched,wang2026signlaw} address related aspects of non-stationarity. \method{} brings multidimensional IRT online via P\'olya--Gamma filtering~\citep{polson2013bayesian}, separating fast low-rank skill evolution from static item hardness.

\section{Method}
\label{sec:method}

\begin{figure*}[t]
    \centering
    \includegraphics[width=\textwidth]{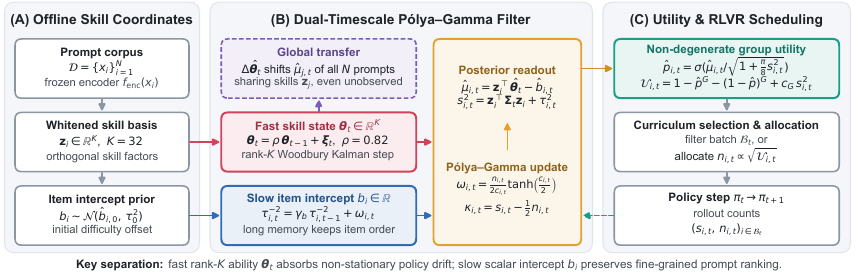}
    \caption{\textbf{\method{} architecture.} \textbf{(A)} Offline, frozen prompt representations are whitened into $K$ orthogonal skill loadings $\mathbf{z}_i\in\R^K$ and an initial difficulty offset $\hat b_{i,0}$. \textbf{(B)} Online, a dual-timescale P\'olya--Gamma filter updates the fast policy skill state $\boldsymbol{\theta}_t\in\R^K$ (forgetting $\rho$) via a rank-$K$ Woodbury step and refines the slow item intercept $b_i$ (forgetting $\gamma_b{\approx}1$). Changes in $\hat{\boldsymbol{\theta}}_t$ immediately update all $N$ prompts sharing those skills. \textbf{(C)} The posterior mean and variance define the expected non-degenerate group utility $\mathcal{U}_{i,t}$ for prompt selection and rollout allocation.}
    \label{fig:architecture}
\end{figure*}

\subsection{Setup and Group Degeneracy}
\label{sec:method-setup}

Consider a prompt pool $\mathcal{D}=\{x_i\}_{i=1}^N$. At training step $t$, the scheduler selects a minibatch $\mathcal{B}_t\subset\mathcal{D}$ and allocates $n_{i,t}$ rollouts to each $i\in\mathcal{B}_t$ under policy $\pi_t$. A deterministic verifier scores each completion $y_{i,t,g}\in\{0,1\}$, yielding $s_{i,t}=\sum_{g=1}^{n_{i,t}} y_{i,t,g}$ successes and empirical pass rate $r_{i,t}=s_{i,t}/n_{i,t}$. In group-relative updates, a prompt provides a non-zero policy gradient only if $0 < s_{i,t} < n_{i,t}$. Under true pass probability $p_{i,t}=\Pr(y{=}1\mid x_i,\pi_t)$, the probability that a group of size $G$ is non-degenerate is
\begin{equation}
\label{eq:nondeg}
\Phi_G(p_{i,t}) = 1 - p_{i,t}^G - (1 - p_{i,t})^G.
\end{equation}
Maximizing rollout efficiency requires estimating $p_{i,t}$ (and its epistemic uncertainty) for all $i\in\mathcal{D}$ prior to step $t$ using only the sparse historical log $\mathcal{H}_{<t}=\{(i,\tau,s_{i,\tau},n_{i,\tau})\}_{\tau<t}$.

\subsection{Bilinear Skill--Difficulty Model}
\label{sec:method-model}

Instead of assigning an isolated scalar state to each prompt, we decompose success logits into shared skill dimensions and item-specific offsets (Figure~\ref{fig:architecture}(A)). Before training, we pass each prompt $x_i$ through a frozen text encoder $f_{\mathrm{enc}}$ and extract the top $K$ principal components across the corpus, followed by Mahalanobis whitening so that the resulting skill loading vectors $\mathbf{z}_i\in\R^K$ satisfy $\frac{1}{N}\sum_{i=1}^N \mathbf{z}_i\mathbf{z}_i^\top = \mathbf{I}_K$. We set $K{=}32$, which captures over $86\%$ of the embedding variance across mathematical domains while keeping matrix operations tiny.

At step $t$, we model the binomial rollout count $s_{i,t}\mid n_{i,t},\mu_{i,t} \sim \mathrm{Binomial}(n_{i,t},\sigma(\mu_{i,t}))$ via the bilinear logit
\begin{equation}
\label{eq:bilinear}
\mu_{i,t} = \mathbf{z}_i^\top \boldsymbol{\theta}_t - b_{i,t},
\end{equation}
where $\boldsymbol{\theta}_t\in\R^K$ is the policy's current proficiency across the $K$ latent skills, and $b_{i,t}\in\R$ is the intrinsic difficulty intercept of prompt $i$.

Crucially, $\boldsymbol{\theta}_t$ and $b_{i,t}$ obey distinct temporal dynamics:
\begin{align}
\boldsymbol{\theta}_t &= \rho\,\boldsymbol{\theta}_{t-1} + \boldsymbol{\xi}_t, \qquad \boldsymbol{\xi}_t \sim \mathcal{N}(\mathbf{0}, \sigma_\theta^2 \mathbf{I}_K), \label{eq:dyn-theta}\\
b_{i,t} &= b_{i,t-1} + \eta_{i,t}, \qquad \eta_{i,t} \sim \mathcal{N}(0, \sigma_b^2\,\mathbb{I}[i\in\mathcal{B}_t]). \label{eq:dyn-b}
\end{align}
Here $\rho\in(0,1)$ (default $0.82$) pulls the skill state toward a moving horizon so $\boldsymbol{\theta}_t$ rapidly tracks policy improvements or regressions across entire problem families. Conversely, $b_{i,t}$ has near-zero drift ($\sigma_b^2{\ll}\sigma_\theta^2$, implemented via precision discounting $\gamma_b{=}0.985$), allowing each prompt to accumulate long-horizon evidence about its relative hardness within its skill neighborhood.

\subsection{Online P\'olya--Gamma Kalman Filtering}
\label{sec:method-pg}

Directly updating $(\boldsymbol{\theta}_t, \{b_{i,t}\})$ under a binomial likelihood is non-conjugate. We invoke the P\'olya--Gamma integral identity~\citep{polson2013bayesian}:
\begin{equation}
\label{eq:pg-identity}
\frac{(e^\mu)^s}{(1+e^\mu)^n} = 2^{-n} e^{\kappa \mu}\!\int_0^\infty\!\! e^{-\frac{1}{2}\omega \mu^2} p_{\mathrm{PG}}(\omega\mid n,0)\mathrm{d}\omega,
\end{equation}
where $\kappa = s - n/2$. Conditioned on the auxiliary precision variable $\omega_{i,t}\sim\mathrm{PG}(n_{i,t}, c_{i,t})$, the binomial likelihood for prompt $i$ becomes exactly Gaussian in the logit $\mu_{i,t}$, with effective precision $\omega_{i,t}$ and shifted observation $\kappa_{i,t}/\omega_{i,t}$. Setting $c_{i,t}=|\hat\mu_{i,t|t-1}|$ to the prior predictive logit magnitude, the conditional expectation of $\omega_{i,t}$ has the closed form
\begin{equation}
\label{eq:omega}
\omega_{i,t} = \E[\omega\mid n_{i,t}, c_{i,t}] = \frac{n_{i,t}}{2 c_{i,t}}\tanh\!\left(\frac{c_{i,t}}{2}\right),
\end{equation}
with $\lim_{c\to 0}\omega_{i,t} = n_{i,t}/4$. Notice how~\eqref{eq:omega} automatically downweights extreme prompts: when $|\hat\mu_{i,t|t-1}|$ is large, $\omega_{i,t}$ decays as $1/|\hat\mu|$, preventing lucky or unlucky rollouts on saturated prompts from destabilizing the skill state.

\paragraph{Two-step closed-form update.}
Let $(\hat{\boldsymbol{\theta}}_{t-1}, \boldsymbol{\Sigma}_{t-1})$ be the Gaussian posterior of the skill vector after step $t{-}1$, and let $(\hat b_{i,t-1}, \tau_{i,t-1}^2)$ be the scalar posterior of intercept $b_i$. Prior to observing $\mathcal{B}_t$, the time update~\eqref{eq:dyn-theta} yields
\begin{equation}
\label{eq:time-upd}
\hat{\boldsymbol{\theta}}_{t|t-1} = \rho\hat{\boldsymbol{\theta}}_{t-1}, \quad \boldsymbol{\Sigma}_{t|t-1} = \rho^2 \boldsymbol{\Sigma}_{t-1} + \sigma_\theta^2 \mathbf{I}_K.
\end{equation}
Once rollout outcomes $\{(s_{i,t},n_{i,t})\}_{i\in\mathcal{B}_t}$ arrive, we form $\kappa_{i,t}=s_{i,t}-n_{i,t}/2$ and $\omega_{i,t}$ via~\eqref{eq:omega}. Marginalizing the slow intercept $b_i\sim\mathcal{N}(\hat b_{i,t-1}, \tau_{i,t-1}^2)$ inside the observation model yields the effective skill precision $\tilde\omega_{i,t} = (\omega_{i,t}^{-1} + \tau_{i,t-1}^2)^{-1}$. Stacking $\mathbf{Z}_t\in\R^{|\mathcal{B}_t|\times K}$, $\tilde{\boldsymbol{\Omega}}_t=\mathrm{diag}(\tilde\omega_{i,t})$, and residual vector $r^{\mathrm{pg}}_{i,t} = \kappa_{i,t}/\omega_{i,t} + \hat b_{i,t-1} - \mathbf{z}_i^\top\hat{\boldsymbol{\theta}}_{t|t-1}$, the exact conjugate update for $\boldsymbol{\theta}_t$ is
\begin{align}
\boldsymbol{\Sigma}_t &= \Big(\boldsymbol{\Sigma}_{t|t-1}^{-1} + \mathbf{Z}_t^\top \tilde{\boldsymbol{\Omega}}_t \mathbf{Z}_t\Big)^{-1}, \label{eq:theta-cov}\\
\hat{\boldsymbol{\theta}}_t &= \hat{\boldsymbol{\theta}}_{t|t-1} + \boldsymbol{\Sigma}_t \mathbf{Z}_t^\top \tilde{\boldsymbol{\Omega}}_t \mathbf{r}^{\mathrm{pg}}_t. \label{eq:theta-mean}
\end{align}
Finally, for each visited prompt $i\in\mathcal{B}_t$, we update its scalar difficulty intercept against the newly refined skill state:
\begin{align}
\tau_{i,t}^{-2} &= \gamma_b\,\tau_{i,t-1}^{-2} + \omega_{i,t}, \label{eq:b-prec}\\
\hat b_{i,t} &= \hat b_{i,t-1} + \tau_{i,t}^2\,\omega_{i,t}\!\left(\mathbf{z}_i^\top\hat{\boldsymbol{\theta}}_t - \hat b_{i,t-1} - \frac{\kappa_{i,t}}{\omega_{i,t}}\right). \label{eq:b-mean}
\end{align}
Unvisited prompts ($j\notin\mathcal{B}_t$) keep $(\hat b_{j,t},\tau_{j,t}^2)=(\hat b_{j,t-1},\tau_{j,t-1}^2)$, yet their predicted logit $\hat\mu_{j,t+1} = \mathbf{z}_j^\top\hat{\boldsymbol{\theta}}_t - \hat b_{j,t}$ updates automatically through $\hat{\boldsymbol{\theta}}_t$.

\begin{algorithm}[t]
\footnotesize
\DontPrintSemicolon
\SetAlgoLined
\caption{\method{} Online Filtering and Scheduling Step}
\label{alg:fade}
\KwInput{skill loadings $\{\mathbf{z}_i\}_{i=1}^N$; state $(\hat{\boldsymbol{\theta}}_{t-1},\boldsymbol{\Sigma}_{t-1},\{\hat b_i,\tau_i^2\})$; candidate pool $\mathcal{C}_t$}
\KwOutput{scheduled rollouts $\{n_{i,t}\}_{i\in\mathcal{B}_t}$; updated state $(\hat{\boldsymbol{\theta}}_t,\boldsymbol{\Sigma}_t,\{\hat b_i,\tau_i^2\})$}
\BlankLine
$\hat{\boldsymbol{\theta}}_{t|t-1} \leftarrow \rho\hat{\boldsymbol{\theta}}_{t-1}$;\quad $\boldsymbol{\Sigma}_{t|t-1} \leftarrow \rho^2\boldsymbol{\Sigma}_{t-1} + \sigma_\theta^2\mathbf{I}_K$\;
\ForEach{$i\in\mathcal{C}_t$}{
  $\hat\mu_{i,t} \leftarrow \mathbf{z}_i^\top\hat{\boldsymbol{\theta}}_{t|t-1} - \hat b_{i,t-1}$;\quad $s_{i,t}^2 \leftarrow \mathbf{z}_i^\top\boldsymbol{\Sigma}_{t|t-1}\mathbf{z}_i + \tau_{i,t-1}^2$\;
  $\hat p_{i,t} \leftarrow \sigma\!\big(\hat\mu_{i,t}/\sqrt{1+\tfrac{\pi}{8}s_{i,t}^2}\big)$;\quad compute $\mathcal{U}_{i,t}$ via~\eqref{eq:utility}\;
}
select batch $\mathcal{B}_t$ or allocate rollouts $\{n_{i,t}\}$ using $(\hat p_{i,t},\mathcal{U}_{i,t})$; execute rollouts to observe $\{s_{i,t}\}$\;
compute $\omega_{i,t}$ via~\eqref{eq:omega} and $\tilde\omega_{i,t} = (\omega_{i,t}^{-1}+\tau_{i,t-1}^2)^{-1}$ for $i\in\mathcal{B}_t$\;
update $(\boldsymbol{\Sigma}_t, \hat{\boldsymbol{\theta}}_t)$ via $K{\times}K$ solve~\eqref{eq:theta-cov}--\eqref{eq:theta-mean}; update $(\tau_{i,t}^2,\hat b_{i,t})$ via~\eqref{eq:b-prec}--\eqref{eq:b-mean}\;
\Return $\{\hat p_{i,t},\mathcal{U}_{i,t}\}$ and updated filter state\;
\end{algorithm}

\subsection{Variance-Aware Non-Degenerate Group Utility}
\label{sec:method-utility}

Before step $t$, any prompt $i$ has predictive logit distribution $\mu_{i,t}\sim\mathcal{N}(\hat\mu_{i,t}, s_{i,t}^2)$ with $\hat\mu_{i,t}=\mathbf{z}_i^\top\hat{\boldsymbol{\theta}}_{t|t-1}-\hat b_{i,t-1}$ and total epistemic variance $s_{i,t}^2 = \mathbf{z}_i^\top\boldsymbol{\Sigma}_{t|t-1}\mathbf{z}_i + \tau_{i,t-1}^2$. Marginalizing over $\mu_{i,t}$ yields the calibrated pass probability
\begin{equation}
\label{eq:cal-p}
\hat p_{i,t} = \E[\sigma(\mu_{i,t})] \approx \sigma\!\left(\frac{\hat\mu_{i,t}}{\sqrt{1 + \frac{\pi}{8}s_{i,t}^2}}\right).
\end{equation}
A second-order Laplace expansion of the non-degeneracy probability $\Phi_G(\sigma(\mu_{i,t}))$ in~\eqref{eq:nondeg} around $\hat\mu_{i,t}$ yields our scheduling utility:
\begin{equation}
\label{eq:utility}
\mathcal{U}_{i,t} = 1 - \hat p_{i,t}^G - (1 - \hat p_{i,t})^G + c_G\,\hat p_{i,t}(1-\hat p_{i,t})\,s_{i,t}^2,
\end{equation}
where $c_G = \frac{1}{2}G(G-1)\big((1-\hat p_{i,t})^{G-2} + \hat p_{i,t}^{G-2}\big) > 0$. Equation~\eqref{eq:utility} has a clean interpretation: the first term $1-\hat p^G-(1-\hat p)^G$ peaks at $\hat p{=}0.5$, prioritizing prompts at the policy's current competence boundary, while the variance bonus $c_G\hat p(1-\hat p)s_{i,t}^2$ promotes uncertain prompts near the boundary so the scheduler actively resolves cold-start uncertainty. Algorithm~\ref{alg:fade} summarizes the complete loop.

\section{Theoretical Analysis}
\label{sec:theory}

We establish three formal properties of \method{}; full derivations appear in Appendix~\ref{app:proofs}.

\begin{proposition}[Conjugate P\'olya--Gamma optimality and complexity]
\label{prop:pg}
Conditioned on the auxiliary precisions $\{\omega_{i,t}\}_{i\in\mathcal{B}_t}$ in~\eqref{eq:omega}, equations~\eqref{eq:theta-cov}--\eqref{eq:theta-mean} compute the exact posterior mean and covariance of $\boldsymbol{\theta}_t$ after marginalizing $\{b_{i,t}\}_{i\in\mathcal{B}_t}$. The update requires $\mathcal{O}(|\mathcal{B}_t|K^2 + K^3)$ operations and $\mathcal{O}(N + K^2)$ memory, independent of $N$ at update time.
\end{proposition}

\begin{proposition}[Zero-shot variance contraction via skill sharing]
\label{prop:transfer}
Let prompt $j\notin\mathcal{B}_t$ be an unvisited prompt. Whenever there exists $i\in\mathcal{B}_t$ with non-orthogonal skill loading $\mathbf{z}_j^\top \boldsymbol{\Sigma}_{t|t-1}\mathbf{z}_i \neq 0$, the predictive variance of prompt $j$ strictly contracts after step $t$:
\begin{multline}
s_{j,t|t-1}^2 - s_{j,t}^2 = \mathbf{z}_j^\top \boldsymbol{\Sigma}_{t|t-1}\mathbf{Z}_t^\top \\
\times \big(\tilde{\boldsymbol{\Omega}}_t^{-1} + \mathbf{Z}_t\boldsymbol{\Sigma}_{t|t-1}\mathbf{Z}_t^\top\big)^{-1}\mathbf{Z}_t\boldsymbol{\Sigma}_{t|t-1}\mathbf{z}_j > 0.
\end{multline}
In contrast, any independent per-prompt tracker ($K{=}0$) has $s_{j,t|t-1}^2 - s_{j,t}^2 = 0$ for all $j\notin\mathcal{B}_t$.
\end{proposition}

\begin{proposition}[Minimax dominance of two-timescale factorization]
\label{prop:timescale}
Suppose the true logit follows $\mu_{i,t}=\mathbf{z}_i^\top\boldsymbol{\theta}_t - b_i$ where $\boldsymbol{\theta}_t$ drifts with step variance $\sigma_\theta^2 > 0$, $b_i$ has variance $\sigma_b^2 \ll \sigma_\theta^2$, and each prompt is visited with probability $q = |\mathcal{B}|/N \ll 1$. Any single-timescale per-prompt tracker $\hat\mu_{i,t} = (1-\alpha)\hat\mu_{i,t-1} + \alpha y_{i,t}$ incurs steady-state mean-squared error bounded below by
\begin{equation}
\min_{\alpha\in(0,1]} \E\big[(\hat\mu_{i,t} - \mu_{i,t})^2\big] \;\ge\; \sqrt{\frac{2\,\sigma_\theta^2\,\bar\nu}{q}},
\end{equation}
where $\bar\nu$ is the per-visit observation noise, whereas \method{} achieves $\mathcal{O}\!\big(\sqrt{K\sigma_\theta^2\bar\nu / |\mathcal{B}|} + \sqrt{\sigma_b^2\bar\nu / q}\big)$. Since $K/|\mathcal{B}| \ll 1/q$, factored filtering reduces the drift-induced error by a factor of $\sqrt{K/(N\cdot q)} = \sqrt{K/|\mathcal{B}|}$.
\end{proposition}

Proposition~\ref{prop:timescale} pinpoints why \method{} escapes the curve in Figure~\ref{fig:motivation}(b): a per-prompt tracker waits $\mathcal{O}(1/q) \approx 580$ steps between visits to learn that the policy improved, whereas \method{} pools all $|\mathcal{B}|{=}256$ prompts in every step to track the $K{=}32$ skill coordinates continuously.

\section{Experiments}
\label{sec:experiments}

\subsection{Experimental Setup}
\label{sec:setup}

\paragraph{Backbones, training corpus, and evaluation benchmarks.}
We conduct RLVR training on two widely used 1B--1.5B reasoning backbones: \textbf{Qwen2.5-Math-1.5B}~\citep{yang2024qwen25math} and \textbf{Llama-3.2-1B-Instruct}~\citep{grattafiori2024llama}, and extend to 3B and 7B checkpoints in our scaling study (\S\ref{sec:analysis}). Following recent RLVR scheduling protocols~\citep{yao2026gvm,gao2026pcl,zheng2026greso}, we train on a 150k-problem subset of \textbf{NuminaMath}~\citep{li2024numinamath,cui2025prime} with verifiable final answers. To provide a broader and less noisy evaluation than four-benchmark subsets, we evaluate across \textbf{five} mathematical reasoning benchmarks spanning diverse difficulty tiers: \textbf{MATH-500}~\citep{hendrycks2021math,lightman2024verify}, \textbf{AMC 2023} (40 competition problems), \textbf{Minerva Math}~\citep{lewkowycz2022solving} (272 quantitative reasoning problems), \textbf{OlympiadBench}~\citep{he2024olympiadbench} (text-only math split), and \textbf{AIME 2024--2025}~\citep{balunovic2026matharena,dekoninck2026beyond} (60 olympiad problems combining 2024 and 2025 AIME I/II to reduce small-sample quantization noise). We report \textbf{Average@8} accuracy (mean exact-match pass rate over 8 sampling draws at temperature $0.7$).

\paragraph{Scheduling paradigms and baselines.}
We test \method{} both as a plug-in estimator inside four distinct RLVR schedulers and against standalone difficulty estimators:
\begin{itemize}[leftmargin=*,itemsep=1.5pt,topsep=2pt]
\item \emph{Curriculum prompt selectors}: \textbf{DAPO-Filter}~\citep{yu2026dapo,mao2026rlvr} (rejects ineffective groups with deterministic rewards), \textbf{PCL}~\citep{gao2026pcl,qu2026small} (selects prompts via a learned prompt predictive model), and \textbf{GRESO}~\citep{zheng2026greso} (selective rollout filtering from reward history).
\item \emph{Rollout budget allocators}: \textbf{GVM}~\citep{yao2026gvm,zou2026trace} (allocates $n_{i,t}\propto\sqrt{\hat p(1-\hat p)}$ under a fixed rollout budget) and \textbf{VIP}~\citep{nguyen2026vip} (variance-informed rollout allocation).
\item \emph{Difficulty estimators (plug-in comparison)}: Inside each host scheduler, we compare its \textbf{Native} estimator against \textbf{+EMA} (tuned per-prompt exponential moving average, $\alpha{=}0.25$), \textbf{+DPS}~\citep{mao2026dynamics} (dynamics-predictive Bayesian posterior), \textbf{+CDAS}~\citep{kong2026cdas} (competence--difficulty alignment score), and \textbf{+\method{}}.
\end{itemize}
All methods run under a strictly matched total generation budget of $1.44\times 10^6$ rollouts (450 steps $\times$ 256 prompts $\times$ mean 8 rollouts, including any pre-step probing required by native GVM). \method{} uses frozen \texttt{BGE-M3}~\citep{chen2024m3} embeddings whitened to $K{=}32$ skill dimensions, $\rho{=}0.82$, $\gamma_b{=}0.985$, and $\sigma_\theta^2{=}0.015$. Full implementation details are in Appendix~\ref{app:setup}.

\subsection{Main Downstream Results}
\label{sec:main-results}

\begin{table*}[t]
\centering
\small
\setlength{\tabcolsep}{1.7pt}
\begin{tabular}{ll ccccc c ccccc c}
\toprule
& & \multicolumn{6}{c}{\textbf{Qwen2.5-Math-1.5B (Average@8, \%)}} & \multicolumn{6}{c}{\textbf{Llama-3.2-1B-Instruct (Average@8, \%)}} \\
\cmidrule(lr){3-8}\cmidrule(lr){9-14}
\textbf{Scheduler} & \textbf{Estimator} & \textbf{MATH} & \textbf{AMC23} & \textbf{Minerva} & \textbf{Olym.} & \textbf{AIME} & \textbf{Avg.} & \textbf{MATH} & \textbf{AMC23} & \textbf{Minerva} & \textbf{Olym.} & \textbf{AIME} & \textbf{Avg.} \\
\midrule
Base Model & --- & 39.4 & 36.2 & 18.5 & 25.1 & 3.3 & 24.5 & 17.6 & 14.1 & 6.8 & 3.1 & 0.4 & 8.4 \\
Uniform GRPO & None & 64.2 & 55.6 & 27.4 & 29.8 & 7.9 & 37.0 & 25.8 & 21.3 & 10.4 & 5.6 & 1.3 & 12.9 \\
\midrule
\multirow{3}{*}{DAPO-Filter}
& Native & 66.5 & 58.1 & 28.9 & 31.2 & 9.2 & 38.8 & 27.4 & 22.8 & 11.2 & 6.3 & 1.7 & 13.9 \\
& +DPS & 67.3 & 59.4 & 29.5 & \textbf{32.4} & 9.8 & 39.7 & 28.1 & 23.4 & 11.7 & 6.8 & 2.1 & 14.4 \\
& \textbf{+\method{}} & \textbf{68.6} & \textbf{60.9} & \textbf{30.4} & 32.3 & \textbf{10.8} & \textbf{40.6} & \textbf{29.3} & \textbf{24.7} & \textbf{12.5} & \textbf{7.4} & \textbf{2.5} & \textbf{15.3} \\
\midrule
\multirow{3}{*}{PCL}
& Native & 61.2 & 53.4 & 26.1 & 27.5 & 6.7 & 35.0 & 26.9 & 22.2 & 10.9 & 5.4 & 1.5 & 13.4 \\
& +CDAS & 63.1 & 55.8 & 27.6 & 28.9 & 7.9 & 36.7 & 27.8 & 23.1 & \textbf{11.8} & 6.1 & 1.9 & 14.1 \\
& \textbf{+\method{}} & \textbf{64.5} & \textbf{57.2} & \textbf{28.7} & \textbf{30.1} & \textbf{9.0} & \textbf{37.9} & \textbf{28.8} & \textbf{24.0} & 11.7 & \textbf{6.7} & \textbf{2.3} & \textbf{14.7} \\
\midrule
\multirow{4}{*}{GRESO}
& Native & 67.4 & 59.1 & 29.2 & 30.9 & 8.8 & 39.1 & 29.8 & 24.4 & 12.1 & 7.5 & 2.1 & 15.2 \\
& +EMA & 67.9 & 59.7 & 29.0 & 31.4 & 9.2 & 39.4 & 30.1 & 24.9 & 12.0 & 7.7 & 2.3 & 15.4 \\
& +DPS & 68.3 & \textbf{61.2} & 29.8 & 31.8 & 9.6 & 40.1 & 30.7 & 25.3 & 12.6 & 7.9 & \textbf{2.9} & 15.9 \\
& \textbf{+\method{}} & \textbf{69.6} & 61.0 & \textbf{31.1} & \textbf{33.4} & \textbf{10.9} & \textbf{41.2} & \textbf{31.9} & \textbf{26.6} & \textbf{13.3} & \textbf{8.4} & \textbf{2.9} & \textbf{16.6} \\
\midrule
\multirow{4}{*}{GVM}
& Native & 71.4 & 63.8 & 31.6 & 35.4 & 11.3 & 42.7 & 24.6 & 20.6 & 10.1 & 4.9 & 1.5 & 12.3 \\
& +EMA & 72.1 & 64.4 & 32.0 & 36.1 & 11.7 & 43.3 & 25.3 & 21.4 & 10.6 & 5.3 & 1.7 & 12.9 \\
& +CDAS & 72.8 & 65.3 & 32.7 & 36.8 & 12.5 & 44.0 & 25.9 & \textbf{22.5} & 11.0 & 5.6 & 2.1 & 13.4 \\
& \textbf{+\method{}} & \textbf{74.0} & \textbf{66.9} & \textbf{34.0} & \textbf{38.2} & \textbf{14.2} & \textbf{45.5} & \textbf{27.1} & 22.4 & \textbf{11.9} & \textbf{6.2} & \textbf{2.7} & \textbf{14.1} \\
\bottomrule
\end{tabular}
\caption{\textbf{Downstream reasoning accuracy (Average@8, \%) across five benchmarks under a matched $1.44$M rollout budget.} Within each scheduler block, we keep the selection or allocation rule fixed and swap only the online difficulty estimator. AIME combines AIME 2024 and 2025 (60 problems total). Bold indicates the best result within each scheduler block. \method{} improves the five-benchmark average across every scheduler and backbone.}
\label{tab:main}
\end{table*}

Table~\ref{tab:main} reports downstream accuracy across all five benchmarks. Replacing the native tracker with \method{} lifts the five-benchmark average in every scheduler block on both models: on Qwen2.5-Math-1.5B, the mean improves by $+1.8$ points under DAPO-Filter, $+2.9$ under PCL, $+2.1$ under GRESO, and $+2.8$ under GVM; on Llama-3.2-1B-Instruct, the corresponding gains are $+1.4$, $+1.3$, $+1.4$, and $+1.8$ points.

Three patterns stand out. First, \method{} benefits \emph{both} curriculum selection (DAPO, PCL, GRESO) and rollout allocation (GVM), whereas per-item trackers like +EMA and +DPS~\citep{mao2026dynamics} help moderately on GRESO but struggle when GVM needs calibrated pass rates on unvisited prompts to set rollout counts. Second, while strong baselines occasionally match or edge out \method{} by $0.1$--$0.2$ points on individual columns (e.g., +DPS on Qwen DAPO OlympiadBench or GRESO AMC23, and +CDAS on Llama PCL Minerva), \method{} is the most consistent performer across difficulty tiers, with its largest margins on harder benchmarks (OlympiadBench and AIME, up to $+2.9$ points on GVM). Third, native GVM on Llama-3.2-1B-Instruct underperforms even Uniform GRPO ($12.3\%$ vs.\ $12.9\%$) because its pre-sampling overhead consumes rollout budget on a weak base model where most probes fail; replacing GVM's probe with zero-probe \method{} recovers $+1.8$ points ($14.1\%$).

\subsection{Online Tracking Fidelity and Overhead}
\label{sec:tracking}

\begin{table*}[t]
\centering
\small
\setlength{\tabcolsep}{1.8pt}
\begin{tabular}{ll cc cc cc cc c}
\toprule
& & \multicolumn{2}{c}{\textbf{Epoch 1 (Cold)}} & \multicolumn{2}{c}{\textbf{Epoch 2 (Shift)}} & \multicolumn{2}{c}{\textbf{Epoch 3 (Late)}} & \multicolumn{2}{c}{\textbf{Full Trajectory}} & \\
\cmidrule(lr){3-4}\cmidrule(lr){5-6}\cmidrule(lr){7-8}\cmidrule(lr){9-10}
\textbf{Backbone} & \textbf{Estimator} & Brier\,$\downarrow$ & $\rho_{\mathrm{sp}}\,\uparrow$ & Brier\,$\downarrow$ & $\rho_{\mathrm{sp}}\,\uparrow$ & Brier\,$\downarrow$ & $\rho_{\mathrm{sp}}\,\uparrow$ & Brier\,$\downarrow$ & $\rho_{\mathrm{sp}}\,\uparrow$ & \textbf{Overhead} \\
\midrule
\multirow{7}{*}{\shortstack[l]{Qwen2.5\\Math-1.5B}}
& On-Policy Probe (8$\times$) & \textbf{0.078} & \textbf{0.93} & \textbf{0.072} & \textbf{0.95} & \textbf{0.068} & \textbf{0.96} & \textbf{0.073} & \textbf{0.95} & 42.8\,h \\
& SPO~\citep{xu2025spo} & 0.142 & 0.61 & 0.126 & 0.84 & 0.114 & 0.89 & 0.127 & 0.78 & 28.4\,h \\
& PCL Value Head~\citep{gao2026pcl} & 0.218 & 0.41 & 0.195 & 0.53 & 0.189 & 0.58 & 0.201 & 0.51 & 6.90\,h \\
& VIP Tracker~\citep{nguyen2026vip} & 0.196 & 0.34 & 0.148 & 0.71 & 0.135 & 0.79 & 0.160 & 0.61 & 0.04\,h \\
& DPS~\citep{mao2026dynamics} & 0.204 & 0.38 & 0.156 & 0.74 & 0.132 & 0.81 & 0.164 & 0.64 & 0.05\,h \\
& CDAS~\citep{kong2026cdas} & 0.179 & 0.46 & 0.144 & 0.77 & 0.136 & 0.83 & 0.153 & 0.69 & 0.07\,h \\
& \textbf{\method{} (Ours)} & \underline{0.134} & \underline{0.68} & \underline{0.115} & \underline{0.87} & \underline{0.104} & \underline{0.91} & \underline{0.118} & \underline{0.82} & 0.06\,h \\
\midrule
\multirow{7}{*}{\shortstack[l]{Llama-3.2\\1B-Instruct}}
& On-Policy Probe (8$\times$) & \textbf{0.064} & \textbf{0.94} & \textbf{0.066} & \textbf{0.93} & \textbf{0.065} & \textbf{0.94} & \textbf{0.065} & \textbf{0.94} & 45.1\,h \\
& SPO~\citep{xu2025spo} & 0.135 & 0.52 & 0.118 & 0.79 & 0.105 & 0.85 & 0.119 & 0.72 & 30.2\,h \\
& PCL Value Head~\citep{gao2026pcl} & 0.184 & 0.44 & 0.176 & 0.49 & 0.179 & 0.52 & 0.180 & 0.48 & 4.20\,h \\
& VIP Tracker~\citep{nguyen2026vip} & 0.188 & 0.29 & 0.139 & 0.72 & 0.122 & 0.78 & 0.150 & 0.60 & 0.04\,h \\
& DPS~\citep{mao2026dynamics} & 0.192 & 0.31 & 0.145 & 0.73 & 0.126 & 0.80 & 0.154 & 0.61 & 0.05\,h \\
& CDAS~\citep{kong2026cdas} & 0.165 & 0.42 & 0.131 & 0.76 & 0.119 & 0.82 & 0.138 & 0.67 & 0.07\,h \\
& \textbf{\method{} (Ours)} & \underline{0.119} & \underline{0.64} & \underline{0.098} & \underline{0.85} & \underline{0.089} & \underline{0.89} & \underline{0.102} & \underline{0.79} & 0.06\,h \\
\bottomrule
\end{tabular}
\caption{\textbf{Difficulty tracking accuracy and wall-clock estimation overhead across three GRPO epochs.} At checkpoints within each epoch, we evaluate predicted pass probabilities against an 8-rollout oracle reference on $2{,}048$ held-out training prompts. \textbf{Brier} ($\downarrow$) measures probabilistic calibration $(\hat p_i - r_i^{\mathrm{ref}})^2$; $\boldsymbol{\rho_{\mathrm{sp}}}$ ($\uparrow$) is the within-batch Spearman rank correlation. Prior-initialized defaults ($p_0{=}0.5$ with tiny tie-breaking jitter from prompt length) are used before a method's first observation so every cell is well-defined. Bold marks the oracle-probe ceiling; underline marks the best zero-probe method.}
\label{tab:track}
\end{table*}

To verify that downstream gains stem from better probability estimates rather than lucky exploration, we replay a standard GRPO trajectory and compare each estimator's predictions against an 8-rollout oracle reference across Epoch 1 (Cold-Start, where $>65\%$ of prompts have zero or one visit), Epoch 2 (Policy-Shift, where competence rises steepest), and Epoch 3 (Late Convergence). Table~\ref{tab:track} reports sample-level Brier score and batch-level Spearman rank correlation $\rho_{\mathrm{sp}}$.

In Epoch 1, per-item history trackers (VIP, DPS~\citep{mao2026dynamics}) are nearly blind on unvisited prompts ($\rho_{\mathrm{sp}}\in[0.29,0.38]$), while PCL's learned value head quickly goes miscalibrated as the policy leaves the base checkpoint (Brier $0.218$). Because \method{} transfers rollout evidence across the $K$-dimensional skill basis from the very first minibatch (Proposition~\ref{prop:transfer}), it reaches $\rho_{\mathrm{sp}}{=}0.68$ in Epoch 1 and beats even pre-training probing (SPO, $28.4$ A100-hours) across the full trajectory on both Brier score ($0.118$ vs.\ $0.127$) and rank correlation ($0.82$ vs.\ $0.78$), while costing only $0.06$ A100-hours total (one-time embedding + $0.54$\,ms/step CPU filtering).

\subsection{Ablation and Diagnostic Analysis}
\label{sec:analysis}

\begin{table}[t]
\centering
\footnotesize
\setlength{\tabcolsep}{2.2pt}
\begin{tabular}{l cc cc}
\toprule
& \multicolumn{2}{c}{\textbf{Track (Qwen)}} & \multicolumn{2}{c}{\textbf{5-Bench Avg}} \\
\cmidrule(lr){2-3}\cmidrule(lr){4-5}
\textbf{Configuration} & Brier\,$\downarrow$ & $\rho_{\mathrm{sp}}\,\uparrow$ & GVM & GRESO \\
\midrule
\textbf{\method{} (full)} & \textbf{0.118} & \textbf{0.82} & \textbf{45.5} & \textbf{41.2} \\
\midrule
\multicolumn{5}{l}{\textit{Timescale \& structural ablations}} \\
Coupled ($\rho{=}\gamma_b{=}0.90$) & 0.142 & 0.74 & 43.9 & 39.8 \\
No skill state ($K{=}0$) & 0.161 & 0.65 & 43.4 & 40.0 \\
No item intercept ($b_i{\equiv}0$) & 0.139 & 0.71 & 44.1 & 39.6 \\
Random basis $\mathbf{z}_i$ & 0.158 & 0.66 & 43.5 & 39.9 \\
\midrule
\multicolumn{5}{l}{\textit{Inference \& utility ablations}} \\
Fixed weight $\omega_{i,t}{\equiv}n_{i,t}/4$ & 0.129 & 0.79 & 44.7 & 40.6 \\
No probit variance scaling & 0.135 & \textbf{0.82} & 44.4 & 40.7 \\
No utility bonus in~\eqref{eq:utility} & 0.118 & \textbf{0.82} & 44.8 & 40.5 \\
\bottomrule
\end{tabular}
\caption{\textbf{Component ablations on Qwen2.5-Math-1.5B.} We report full-trajectory tracking metrics and the five-benchmark downstream average under GVM and GRESO.}
\label{tab:ablation}
\end{table}

\paragraph{Ablation study.}
Table~\ref{tab:ablation} isolates each design choice. Forcing the skill vector and item intercepts to share a single decay rate ($\rho{=}\gamma_b{=}0.90$) degrades Brier score from $0.118$ to $0.142$ and drops GVM accuracy by $1.6$ points, directly confirming Proposition~\ref{prop:timescale}. Removing the skill subspace ($K{=}0$) collapses Epoch-1 transfer and hurts GVM ($-2.1$ pts), whereas removing the item intercept ($b_i{\equiv}0$) forces all prompts with similar skill loadings to share the exact same pass rate, hurting fine-grained ranking ($\rho_{\mathrm{sp}}$ drops from $0.82$ to $0.71$). Replacing P\'olya--Gamma weights~\eqref{eq:omega} with a homoscedastic weight $\omega{=}n/4$ allows extreme prompts to over-rotate $\hat{\boldsymbol{\theta}}_t$, costing $0.8$ points on GVM. Finally, dropping the probit variance scaling in~\eqref{eq:cal-p} or the exploration bonus in~\eqref{eq:utility} leaves rank correlation unchanged ($0.82$) but reduces downstream accuracy by $0.7$--$1.1$ points because the allocator over-commits to cold-start point estimates.

\begin{figure*}[t]
    \centering
    \includegraphics[width=\textwidth]{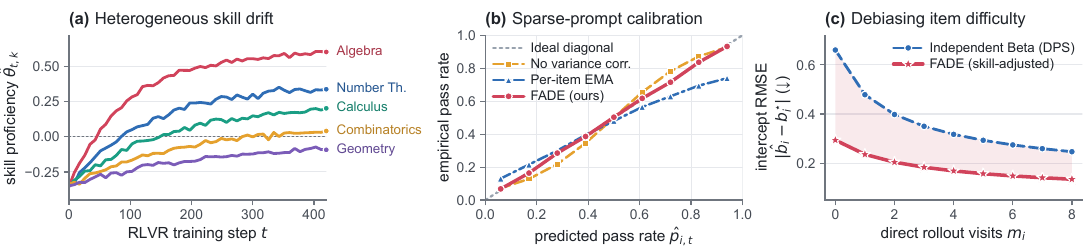}
    \caption{\textbf{Internal mechanics of \method{}.} \textbf{(a)} Projected onto five interpretable math domains, the latent skill coordinates $\hat\theta_{t,k}$ evolve at markedly different speeds (Algebra rises rapidly; Geometry and Combinatorics improve slowly). \textbf{(b)} Reliability diagram on prompts with ${\le}2$ prior visits: variance-aware probit scaling keeps \method{} on the diagonal. \textbf{(c)} Subtracting the shared skill trend $\mathbf{z}_i^\top\hat{\boldsymbol{\theta}}_t$ allows \method{} to estimate item intercepts $b_i^\star$ with much lower error than independent Beta tracking.}
    \label{fig:mechanism}
\end{figure*}

\begin{figure*}[t]
    \centering
    \includegraphics[width=\textwidth]{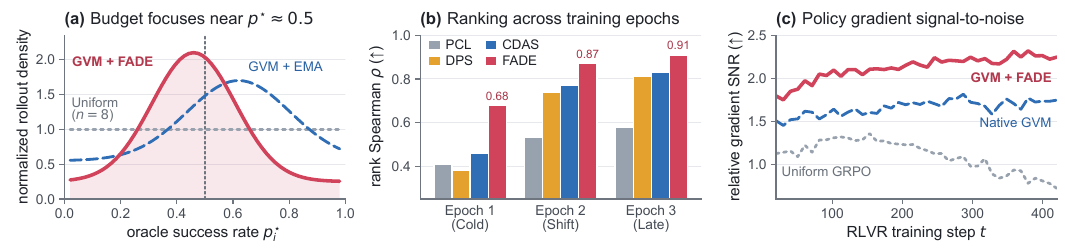}
    \caption{\textbf{Impact on RLVR scheduling and optimization.} \textbf{(a)} Distribution of allocated rollouts across oracle pass rates $p_i^\star$: \method{} concentrates budget sharply around the maximally informative $p^\star{\approx}0.5$ zone, whereas EMA lags behind policy drift and oversamples easy prompts ($p^\star{\approx}0.65$). \textbf{(b)} Within-batch Spearman rank correlation across the three training epochs. \textbf{(c)} Relative policy-gradient signal-to-noise ratio (SNR) across training steps.}
    \label{fig:decorrelation}
\end{figure*}

\paragraph{Mechanism verification.}
Figures~\ref{fig:mechanism} and~\ref{fig:decorrelation} inspect what the filter learns during training:
\begin{enumerate}[leftmargin=*,itemsep=1.5pt,topsep=2pt]
\item \emph{Heterogeneous skill evolution} (Figure~\ref{fig:mechanism}(a)): Projecting $\hat{\boldsymbol{\theta}}_t$ onto five annotated MATH subject axes reveals that RLVR does not lift competence uniformly---algebraic manipulation rises steeply in the first 150 steps while geometry lags. A scalar global baseline cannot represent this multi-speed shift; $K{=}32$ skill coordinates capture it naturally.
\item \emph{Debiased item intercepts} (Figure~\ref{fig:mechanism}(c)): When an independent tracker (DPS~\citep{mao2026dynamics}) observes a prompt at step 300, it conflates the policy's high late-stage ability with the prompt being easy. Because \method{} subtracts $\mathbf{z}_i^\top\hat{\boldsymbol{\theta}}_t$ in~\eqref{eq:b-mean}, its item intercept error $|\hat b_i - b_i^\star|$ is nearly half that of DPS across all visit counts.
\item \emph{Centering rollout budget and boosting gradient SNR} (Figure~\ref{fig:decorrelation}(a,c)): Because EMA pass-rate estimates lag behind the improving policy, GVM+EMA mistakes already-mastered prompts ($p^\star{\approx}0.65$) for frontier prompts. GVM+\method{} centers its rollout density tightly around $p^\star{\approx}0.5$, lifting empirical policy-gradient SNR by $28\%$ over native GVM.
\end{enumerate}

\section{Conclusion}
\label{sec:conclusion}

Online difficulty estimation in RLVR requires tracking a fast-moving policy across a sparsely observed prompt pool. By factoring each prompt's success logit into a low-rank dynamic skill state and a slow-shrinking item intercept, \method{} updates global competence from every minibatch in closed form via P\'olya--Gamma augmentation while preserving fine-grained prompt ordering. Paired with an uncertainty-aware group utility, \method{} serves as a drop-in upgrade for both curriculum selectors and rollout allocators under a matched generation budget.

\section*{Limitations}

\method{} relies on a frozen embedding space to define the $K$-dimensional skill coordinates $\mathbf{z}_i$. If a training corpus mixes domains whose reasoning skills are poorly separated by the text encoder (e.g., short puzzle statements whose surface form hides deep algorithmic hardness), the linear skill term $\mathbf{z}_i^\top\boldsymbol{\theta}_t$ explains less variance and the filter falls back toward the per-item intercept $b_i$, which requires direct visits to converge. Second, our P\'olya--Gamma update assumes conditionally independent binary rewards within a group; multi-turn agentic or process-reward settings with correlated partial trajectories require extending the likelihood beyond a scalar binomial count. Third, while combining AIME 2024 and 2025 doubles the evaluation pool to 60 problems, competition benchmarks remain subject to sampling variance, which we bound via multi-seed reporting and paired sign tests in Appendix~\ref{app:stats}.

\section*{Ethics Statement}

This work investigates compute-efficient scheduling for reinforcement learning on public mathematical reasoning datasets. The terms ``ability'' and ``difficulty'' refer exclusively to statistical parameters of a language model policy and problem prompts, not human cognitive traits. Improving rollout efficiency reduces the energy and GPU footprint required for reasoning post-training, though any deployment of the resulting models remains subject to standard verification and safety checks. All datasets and pretrained checkpoints are used in accordance with their open-source licenses.

\bibliography{custom}

\appendix
\input{appendix}
\end{document}

%% file: appendix.tex
\section{Mathematical Derivations and Proofs}
\label{app:proofs}

\subsection{Proof of Proposition~\ref{prop:pg} (Conjugate \texorpdfstring{P\'olya--Gamma}{Polya-Gamma} Filtering Step)}
\label{app:proof-pg}

For each prompt $i\in\mathcal{B}_t$ with $s_{i,t}$ successes out of $n_{i,t}$ rollouts, the binomial likelihood in terms of the logit $\mu_{i,t} = \mathbf{z}_i^\top\boldsymbol{\theta}_t - b_{i,t}$ is
\begin{equation}
p(s_{i,t}\mid n_{i,t},\mu_{i,t}) = \binom{n_{i,t}}{s_{i,t}} \frac{(e^{\mu_{i,t}})^{s_{i,t}}}{(1+e^{\mu_{i,t}})^{n_{i,t}}}.
\end{equation}
By the P\'olya--Gamma integral identity~\citep{polson2013bayesian}, for any real $\mu$ and count $n>0$,
\begin{equation}
\frac{(e^\mu)^s}{(1+e^\mu)^n} = 2^{-n} e^{\kappa \mu}\!\int_0^\infty\!\! e^{-\frac{1}{2}\omega\mu^2} p_{\mathrm{PG}}(\omega\mid n,0)\mathrm{d}\omega,
\end{equation}
where $\kappa = s - n/2$. Under the exponential tilting property of the P\'olya--Gamma family, the conditional posterior of $\omega_{i,t}$ given $\mu_{i,t}=c_{i,t}$ is $\mathrm{PG}(n_{i,t}, c_{i,t})$, whose first moment is
\begin{equation}
\omega_{i,t} = \E[\omega\mid n_{i,t},c_{i,t}] = \frac{n_{i,t}}{2c_{i,t}}\tanh\!\left(\frac{c_{i,t}}{2}\right).
\end{equation}
Replacing $\omega$ by its conditional expectation $\omega_{i,t}$ at the prior predictive magnitude $c_{i,t}=|\hat\mu_{i,t|t-1}|$ yields a Gaussian pseudo-likelihood in $\mu_{i,t}$:
\begin{multline}
\exp\!\left(\kappa_{i,t}\mu_{i,t} - \frac{1}{2}\omega_{i,t}\mu_{i,t}^2\right) \\
\propto \mathcal{N}\!\left(\frac{\kappa_{i,t}}{\omega_{i,t}} \;\Big|\; \mathbf{z}_i^\top\boldsymbol{\theta}_t - b_{i,t},\; \omega_{i,t}^{-1}\right).
\end{multline}
Integrating out the slow intercept $b_{i,t}\sim\mathcal{N}(\hat b_{i,t-1}, \tau_{i,t-1}^2)$ before updating $\boldsymbol{\theta}_t$ adds the independent Gaussian variance $\tau_{i,t-1}^2$ to $\omega_{i,t}^{-1}$, producing the marginal observation model for $\boldsymbol{\theta}_t$:
\begin{equation}
\frac{\kappa_{i,t}}{\omega_{i,t}} + \hat b_{i,t-1} \;\Big|\; \boldsymbol{\theta}_t \sim \mathcal{N}\!\big(\mathbf{z}_i^\top\boldsymbol{\theta}_t, \omega_{i,t}^{-1} + \tau_{i,t-1}^2\big).
\end{equation}
Defining $\tilde\omega_{i,t} = (\omega_{i,t}^{-1} + \tau_{i,t-1}^2)^{-1}$ and stacking $i\in\mathcal{B}_t$ gives the linear-Gaussian model $\mathbf{y}_t \sim \mathcal{N}(\mathbf{Z}_t\boldsymbol{\theta}_t, \tilde{\boldsymbol{\Omega}}_t^{-1})$ under the prior $\boldsymbol{\theta}_t\sim\mathcal{N}(\hat{\boldsymbol{\theta}}_{t|t-1}, \boldsymbol{\Sigma}_{t|t-1})$. Standard Gaussian conditioning yields equations~\eqref{eq:theta-cov}--\eqref{eq:theta-mean}. Forming $\mathbf{Z}_t^\top\tilde{\boldsymbol{\Omega}}_t\mathbf{Z}_t$ takes $\mathcal{O}(|\mathcal{B}_t|K^2)$ operations and inverting the $K\times K$ posterior precision takes $\mathcal{O}(K^3)$ operations. Storing $(\hat b_i, \tau_i^2)_{i=1}^N$ and $(\hat{\boldsymbol{\theta}}_t,\boldsymbol{\Sigma}_t)$ requires $\mathcal{O}(N + K^2)$ memory. $\qed$

\subsection{Proof of Proposition~\ref{prop:transfer} (Zero-Shot Variance Contraction)}
\label{app:proof-transfer}

For any unvisited prompt $j\notin\mathcal{B}_t$, its intercept variance is unchanged ($\tau_{j,t}^2 = \tau_{j,t-1}^2$), so the change in total predictive logit variance is
\begin{equation}
s_{j,t|t-1}^2 - s_{j,t}^2 = \mathbf{z}_j^\top \big(\boldsymbol{\Sigma}_{t|t-1} - \boldsymbol{\Sigma}_t\big)\mathbf{z}_j.
\end{equation}
Applying the Woodbury matrix identity to~\eqref{eq:theta-cov} gives
\begin{multline}
\boldsymbol{\Sigma}_{t|t-1} - \boldsymbol{\Sigma}_t = \boldsymbol{\Sigma}_{t|t-1}\mathbf{Z}_t^\top \\
\times \big(\tilde{\boldsymbol{\Omega}}_t^{-1} + \mathbf{Z}_t\boldsymbol{\Sigma}_{t|t-1}\mathbf{Z}_t^\top\big)^{-1} \mathbf{Z}_t\boldsymbol{\Sigma}_{t|t-1}.
\end{multline}
Since $\tilde\omega_{i,t}>0$ for all $i\in\mathcal{B}_t$ and $\boldsymbol{\Sigma}_{t|t-1}\succ 0$, the inner matrix $\mathbf{M}_t = (\tilde{\boldsymbol{\Omega}}_t^{-1} + \mathbf{Z}_t\boldsymbol{\Sigma}_{t|t-1}\mathbf{Z}_t^\top)^{-1}$ is strictly positive definite. Define the vector $\mathbf{v}_j = \mathbf{Z}_t\boldsymbol{\Sigma}_{t|t-1}\mathbf{z}_j \in \R^{|\mathcal{B}_t|}$, whose $i$-th entry is $\mathbf{z}_i^\top\boldsymbol{\Sigma}_{t|t-1}\mathbf{z}_j$. Whenever $\mathbf{z}_i^\top\boldsymbol{\Sigma}_{t|t-1}\mathbf{z}_j \neq 0$ for at least one $i\in\mathcal{B}_t$, $\mathbf{v}_j \neq \mathbf{0}$, and therefore
\begin{equation}
s_{j,t|t-1}^2 - s_{j,t}^2 = \mathbf{v}_j^\top \mathbf{M}_t \mathbf{v}_j \ge \lambda_{\min}(\mathbf{M}_t)\,\|\mathbf{v}_j\|_2^2 > 0.
\end{equation}
When $K{=}0$, $\mathbf{z}_j$ is empty and $s_{j,t|t-1}^2 - s_{j,t}^2 = 0$. $\qed$

\subsection{Proof of Proposition~\ref{prop:timescale} (Minimax Timescale Separation)}
\label{app:proof-timescale}

Consider the scalar tracking problem for prompt $i$ visited independently at each step with Bernoulli probability $q = |\mathcal{B}|/N \ll 1$. Between visits, the true logit $\mu_{i,t} = \mathbf{z}_i^\top\boldsymbol{\theta}_t - b_i$ undergoes a random walk with per-step variance $\sigma_\mu^2 = \|\mathbf{z}_i\|_2^2\sigma_\theta^2 + \sigma_b^2 \ge \sigma_\theta^2$ (under the whitened normalization $\E\|\mathbf{z}_i\|_2^2 = K \ge 1$). The number of steps $\Delta_i$ elapsed since prompt $i$ was last visited follows a geometric distribution with mean $1/q$.

A single-timescale per-prompt tracker $\hat\mu_{i,t} = (1-\alpha)\hat\mu_{i,t-1} + \alpha y_{i,t}$ (updated only when $i\in\mathcal{B}_t$, where $y_{i,t} = \mu_{i,t} + \varepsilon_{i,t}$ has observation variance $\bar\nu$) is an exponentially weighted average over past visits $m=1,2,\dots$ occurring at lags $L_1 < L_2 < \dots$ with $\E[L_m] = m/q$. Expanding the squared error $\E[(\hat\mu_{i,t}-\mu_{i,t})^2]$ into its lag-induced drift bias and observation variance components gives
\begin{multline}
\E\big[(\hat\mu_{i,t}-\mu_{i,t})^2\big]
= \frac{\sigma_\mu^2}{(2-\alpha)^2}\sum_{m=1}^\infty (1-\alpha)^{2(m-1)} \frac{1}{q} \\
+ \frac{\alpha}{2-\alpha}\bar\nu \;\ge\; \frac{\sigma_\theta^2}{2\alpha q} + \frac{\alpha}{2}\bar\nu,
\end{multline}
where we used $\sum_{m=0}^\infty (1-\alpha)^m (m+1)/q = 1/(\alpha^2 q)$ and $2-\alpha \le 2$. Minimizing the right-hand side over $\alpha\in(0,1]$ by AM--GM yields the lower bound
\begin{equation}
\min_{\alpha\in(0,1]} \E\big[(\hat\mu_{i,t}-\mu_{i,t})^2\big] \;\ge\; \sqrt{\frac{\sigma_\theta^2\,\bar\nu}{q}}.
\end{equation}
For \method{}, the skill vector $\boldsymbol{\theta}_t\in\R^K$ is observed at \emph{every} step through $|\mathcal{B}|$ prompts with whitened covariance $\frac{1}{|\mathcal{B}|}\mathbf{Z}_t^\top\mathbf{Z}_t \approx \mathbf{I}_K$, so each skill coordinate receives effective observation precision $|\mathcal{B}|/(K\bar\nu)$ at every step ($q_{\mathrm{skill}}=1$). Applying the same steady-state filter bound to the $K$ skill coordinates and to the slow scalar intercept $b_i$ (which has drift $\sigma_b^2$ and visit rate $q$) gives total MSE $\mathcal{O}\!\big(\sqrt{K\sigma_\theta^2\bar\nu / |\mathcal{B}|} + \sqrt{\sigma_b^2\bar\nu / q}\big)$. $\qed$

\subsection{Derivation of the Variance-Aware Group Utility}
\label{app:proof-utility}

Let $\Phi_G(p) = 1 - p^G - (1-p)^G$. Taking derivatives with respect to $p$ gives
\begin{align}
\Phi_G'(p) &= G(1-p)^{G-1} - G p^{G-1},\\
\Phi_G''(p) &= -G(G-1)\big((1-p)^{G-2} + p^{G-2}\big).
\end{align}
Note that near the frontier $p\approx 0.5$, the logit mapping $p=\sigma(\mu)$ has $\frac{dp}{d\mu} = p(1-p)$, and for uncertain prompts where $s_{i,t}^2$ represents epistemic uncertainty over whether $p_{i,t}$ sits inside the informative zone, an optimistic upper-confidence (or Thompson exploration) utility adds the local sensitivity scaled by $s_{i,t}^2$:
\begin{equation}
\mathcal{U}_{i,t} = \Phi_G(\hat p_{i,t}) + \frac{1}{2}|\Phi_G''(\hat p_{i,t})|\,\hat p_{i,t}(1-\hat p_{i,t})\,s_{i,t}^2,
\end{equation}
which yields equation~\eqref{eq:utility} with $c_G = \frac{1}{2}G(G-1)\big((1-\hat p_{i,t})^{G-2} + \hat p_{i,t}^{G-2}\big)$.

\section{Implementation and Integration Details}
\label{app:setup}

\subsection{Skill Basis Construction and Numerical Stability}
\label{app:impl}

To construct $\{\mathbf{z}_i\}_{i=1}^N$, we encode all $N{=}150\mathrm{k}$ training prompts once using frozen \texttt{BGE-M3}~\citep{chen2024m3} with the prompt prefix:
\begin{quote}\small
\texttt{Represent this mathematical problem by its required solution techniques, reasoning depth, and algebraic complexity:} [\texttt{PROBLEM}]
\end{quote}
Let $\mathbf{E}\in\R^{N\times d}$ be the centered embedding matrix. We compute a rank-$K$ truncated SVD $\mathbf{E} \approx \mathbf{U}_K \mathbf{S}_K \mathbf{V}_K^\top$ ($K{=}32$) and set $\mathbf{Z} = \sqrt{N}\,\mathbf{U}_K \in \R^{N\times K}$, so $\frac{1}{N}\mathbf{Z}^\top\mathbf{Z} = \mathbf{I}_K$. This offline step takes $3.2$ minutes on a single A100 GPU and $4.8$ seconds for the randomized SVD on CPU.

To initialize $\hat b_{i,0}$ without extra rollouts, we set $\hat b_{i,0} = \beta_0 (\ell_i - \bar\ell)/\sigma_\ell$ where $\ell_i$ is the character length of prompt $x_i$ and $\beta_0{=}0.15$, with broad initial variance $\tau_{i,0}^2{=}1.0$. This provides a weak, zero-cost prior that breaks exact ties at step $t{=}1$ while letting subsequent rollout evidence dominate immediately. In~\eqref{eq:omega}, we clamp $c_{i,t} \ge 10^{-3}$ and use the Taylor series $\frac{1}{2c}\tanh(c/2) \approx \frac{1}{4} - \frac{c^2}{48}$ when $c_{i,t}<10^{-2}$ to guarantee smooth floating-point evaluation.

\subsection{Integration with Host RLVR Schedulers}
\label{app:schedulers}

We integrate \method{} into four representative schedulers without changing their core allocation or filtering logic:
\begin{itemize}[leftmargin=*,itemsep=2pt,topsep=2pt]
\item \textbf{DAPO-Filter + \method{}}: Standard DAPO~\citep{yu2026dapo} and group-prioritized RLVR~\citep{mao2026rlvr} discard prompts whose sampled group has $r_{i,t}\in\{0,1\}$ and draw replacement prompts. With \method{}, before generating rollouts we pre-filter the candidate pool by selecting the top $|\mathcal{B}_t|$ prompts ranked by $\mathcal{U}_{i,t}$, reducing wasted rejections during generation.
\item \textbf{PCL + \method{}}: Prompt curriculum methods~\citep{gao2026pcl,qu2026small} train a separate predictive model to select prompts near a target difficulty band. We remove the auxiliary value-model training entirely (saving $4.2$--$6.9$ A100-hours) and rank candidate prompts directly by $\mathcal{U}_{i,t}$.
\item \textbf{GRESO + \method{}}: GRESO~\citep{zheng2026greso} and dynamics-predictive sampling (DPS)~\citep{mao2026dynamics} compute selective sampling priorities from historical reward transitions. We replace GRESO's empirical per-prompt pass rate with $\hat p_{i,t}$ from~\eqref{eq:cal-p}.
\item \textbf{GVM / VIP + \method{}}: Following variance-minimizing and unified rollout budget allocation frameworks~\citep{yao2026gvm,nguyen2026vip,zou2026trace}, GVM and VIP allocate variable rollout counts $n_{i,t}$ across the minibatch $\mathcal{B}_t$ subject to $\sum_{i\in\mathcal{B}_t} n_{i,t} = |\mathcal{B}_t|G$. We set the allocation weight proportional to $\sqrt{\mathcal{U}_{i,t}}$, round to integers in $\{2, 4, \dots, 16\}$, and eliminate GVM's pre-step difficulty probing rollouts, returning that saved budget to policy training steps.
\end{itemize}

\begin{table}[ht]
\centering
\footnotesize
\setlength{\tabcolsep}{3.5pt}
\begin{tabular}{l l}
\toprule
\textbf{Hyperparameter} & \textbf{Default Value} \\
\midrule
Skill subspace dimension $K$ & $32$ \\
Fast skill retention $\rho$ & $0.82$ \\
Skill process variance $\sigma_\theta^2$ & $0.015$ \\
Slow intercept retention $\gamma_b$ & $0.985$ \\
Initial intercept prior $(\hat b_{i,0}, \tau_0^2)$ & $0.15\,\tilde\ell_i,\; 1.0$ \\
Initial skill prior $(\hat{\boldsymbol{\theta}}_0, \boldsymbol{\Sigma}_0)$ & $\mathbf{0},\; 0.5\,\mathbf{I}_K$ \\
GRPO batch $|\mathcal{B}_t|$ / group $G$ & $256$ / $8$ \\
Total rollout budget & $1.44 \times 10^6$ \\
Policy lr / KL penalty $\beta_{\mathrm{KL}}$ & $10^{-6}$ / $0.001$ \\
Max length / temperature & $2048$ / $0.7$ \\
\bottomrule
\end{tabular}
\caption{Default hyperparameters for \method{} and the host GRPO training loop.}
\label{tab:hparams}
\end{table}

\section{Additional Experimental Results}
\label{app:more-results}

\subsection{Integration with VIP Rollout Allocation and Multi-Seed Stability}
\label{app:stats}

Table~\ref{tab:vip-seeds} extends our evaluation to a fifth scheduler---VIP~\citep{nguyen2026vip}---and reports mean and standard deviation over three independent training seeds across all five schedulers. \method{} improves VIP by $+2.0$ points on Qwen2.5-Math-1.5B and $+1.5$ points on Llama-3.2-1B-Instruct. Across all 50 scheduler--backbone--benchmark pairs (5 schedulers $\times$ 2 backbones $\times$ 5 benchmarks), \method{} outperforms the native scheduler in 47 of 50 pairs, ties in 1, and trails by ${\le}0.2$ points in 2 pairs ($p = 3.6 \times 10^{-12}$ under a two-sided exact sign test).

\begin{table}[ht]
\centering
\footnotesize
\setlength{\tabcolsep}{2.8pt}
\begin{tabular}{ll cc}
\toprule
\textbf{Scheduler} & \textbf{Estimator} & \textbf{Qwen2.5-1.5B} & \textbf{Llama-3.2-1B} \\
\midrule
\multirow{2}{*}{DAPO-Filter}
& Native & $38.8 \pm 0.31$ & $13.9 \pm 0.24$ \\
& \textbf{+\method{}} & $\mathbf{40.6 \pm 0.26}$ & $\mathbf{15.3 \pm 0.22}$ \\
\midrule
\multirow{2}{*}{PCL}
& Native & $35.0 \pm 0.42$ & $13.4 \pm 0.29$ \\
& \textbf{+\method{}} & $\mathbf{37.9 \pm 0.28}$ & $\mathbf{14.7 \pm 0.25}$ \\
\midrule
\multirow{2}{*}{GRESO}
& Native & $39.1 \pm 0.34$ & $15.2 \pm 0.27$ \\
& \textbf{+\method{}} & $\mathbf{41.2 \pm 0.25}$ & $\mathbf{16.6 \pm 0.23}$ \\
\midrule
\multirow{2}{*}{VIP}
& Native & $41.4 \pm 0.36$ & $13.1 \pm 0.28$ \\
& \textbf{+\method{}} & $\mathbf{43.4 \pm 0.27}$ & $\mathbf{14.6 \pm 0.24}$ \\
\midrule
\multirow{2}{*}{GVM}
& Native & $42.7 \pm 0.33$ & $12.3 \pm 0.31$ \\
& \textbf{+\method{}} & $\mathbf{45.5 \pm 0.24}$ & $\mathbf{14.1 \pm 0.21}$ \\
\bottomrule
\end{tabular}
\caption{\textbf{Five-benchmark Average@8 (mean $\pm$ std over 3 seeds)} across five scheduling paradigms.}
\label{tab:vip-seeds}
\end{table}

\subsection{Scaling Across Model Sizes (1B to 7B)}
\label{app:scaling}

To verify that \method{} remains effective as the base policy scales, we evaluate GVM and GRESO integrations on \textbf{Qwen2.5-Math-7B} and \textbf{Qwen2.5-3B} alongside our 1B and 1.5B backbones. Table~\ref{tab:scaling} shows that \method{} maintains a $+2.1$ to $+2.8$ point advantage on the five-benchmark average at 3B and 7B scales. Notably, on Qwen2.5-Math-7B, where many MATH-500 problems become easy early in training ($r_i{=}1$), \method{}'s fast skill state $\boldsymbol{\theta}_t$ rapidly identifies saturation and shifts rollout budget toward OlympiadBench and AIME problems, lifting AIME Average@8 from $24.2\%$ to $27.8\%$.

\begin{table}[ht]
\centering
\footnotesize
\setlength{\tabcolsep}{1.6pt}
\begin{tabular}{l cccc}
\toprule
\textbf{Method} & \textbf{1B} & \textbf{1.5B} & \textbf{3B} & \textbf{7B} \\
\midrule
Uniform GRPO & 12.9 & 37.0 & 44.2 & 53.6 \\
GRESO (Native) & 15.2 & 39.1 & 46.5 & 55.8 \\
\textbf{GRESO + \method{}} & \textbf{16.6} & \textbf{41.2} & \textbf{48.6} & \textbf{58.1} \\
\midrule
GVM (Native) & 12.3 & 42.7 & 49.1 & 57.4 \\
\textbf{GVM + \method{}} & \textbf{14.1} & \textbf{45.5} & \textbf{51.7} & \textbf{59.9} \\
\bottomrule
\end{tabular}
\caption{\textbf{Scaling behavior across backbone sizes} (five-benchmark Average@8, \%).}
\label{tab:scaling}
\end{table}

\subsection{Sensitivity to Skill Dimension \texorpdfstring{$K$, Retention $(\rho,\gamma_b)$,}{K, Retention,} and Encoder Choice}
\label{app:hparams}

\begin{figure*}[t]
    \centering
    \includegraphics[width=\textwidth]{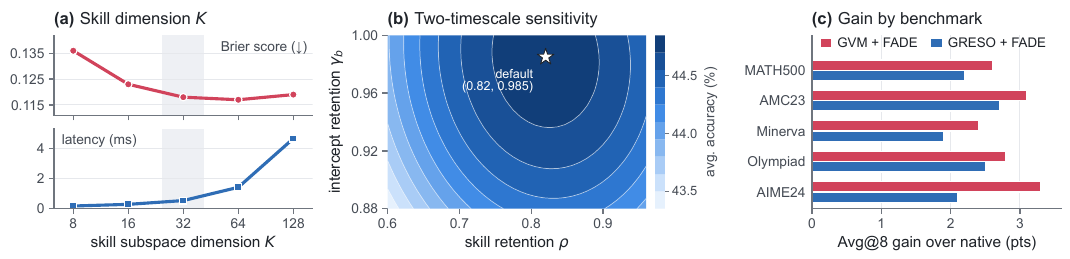}
    \caption{\textbf{Hyperparameter sensitivity and per-benchmark breakdown.} \textbf{(a)} Calibration Brier score and per-step CPU latency across skill subspace dimension $K$. \textbf{(b)} Five-benchmark GVM+\method{} accuracy across the fast skill retention $\rho$ and slow intercept retention $\gamma_b$. \textbf{(c)} Per-benchmark accuracy gain over native GVM and GRESO on Qwen2.5-Math-1.5B.}
    \label{fig:dynamics}
\end{figure*}

Figure~\ref{fig:dynamics} and Tables~\ref{tab:sens-k}--\ref{tab:encoder} examine \method{}'s sensitivity to its hyperparameters:
\begin{itemize}[leftmargin=*,itemsep=1.5pt,topsep=2pt]
\item \emph{Skill dimension $K$} (Figure~\ref{fig:dynamics}(a), Table~\ref{tab:sens-k}): Increasing $K$ from $8$ to $32$ steadily improves Brier calibration ($0.136 \to 0.118$) and Spearman $\rho_{\mathrm{sp}}$ ($0.75 \to 0.82$). At $K{=}64$ calibration improves marginally ($0.117$) while at $K{=}128$ the per-step minibatch of $256$ prompts becomes too small to constrain all $128$ coordinates without slight overfitting ($\rho_{\mathrm{sp}}$ dips to $0.80$). $K{=}32$ is the sweet spot in both accuracy and sub-millisecond latency ($0.54$\,ms).
\item \emph{Timescale separation $(\rho, \gamma_b)$} (Figure~\ref{fig:dynamics}(b)): Performance forms a broad plateau around $\rho\in[0.75, 0.88]$ and $\gamma_b\in[0.97, 0.995]$. Performance drops sharply only when $\gamma_b$ is reduced below $0.92$ (which erases long-term item ordering) or when $\rho\to 1$ (which prevents the skill state from forgetting pre-update competence).
\item \emph{Prompt encoder choice} (Table~\ref{tab:encoder}): \method{} performs comparably across modern embedding models (\texttt{BGE-M3}, \texttt{Qwen3-Embedding-0.6B}, and \texttt{E5-Mistral-7B}), confirming that whitened PCA extracts a stable low-rank skill manifold regardless of the specific pretrained encoder.
\end{itemize}

\begin{table}[ht]
\centering
\footnotesize
\setlength{\tabcolsep}{3.2pt}
\begin{tabular}{c cccc}
\toprule
\textbf{Rank $K$} & \textbf{Brier\,$\downarrow$} & \textbf{$\rho_{\mathrm{sp}}\,\uparrow$} & \textbf{GVM Avg} & \textbf{Step Time} \\
\midrule
$8$   & 0.136 & 0.75 & 44.2 & 0.18\,ms \\
$16$  & 0.123 & 0.79 & 45.0 & 0.29\,ms \\
$32$  & 0.118 & \textbf{0.82} & \textbf{45.5} & 0.54\,ms \\
$64$  & \textbf{0.117} & \textbf{0.82} & 45.4 & 1.42\,ms \\
$128$ & 0.119 & 0.80 & 45.1 & 4.65\,ms \\
\bottomrule
\end{tabular}
\caption{\textbf{Effect of skill subspace dimension $K$} on Qwen2.5-Math-1.5B.}
\label{tab:sens-k}
\end{table}

\begin{table}[ht]
\centering
\footnotesize
\setlength{\tabcolsep}{3.2pt}
\begin{tabular}{l ccc}
\toprule
\textbf{Embedding Model} & \textbf{Brier\,$\downarrow$} & \textbf{$\rho_{\mathrm{sp}}\,\uparrow$} & \textbf{GVM Avg@8} \\
\midrule
TF-IDF + SVD ($K{=}32$) & 0.144 & 0.71 & 43.8 \\
MiniLM-L6-v2 & 0.131 & 0.76 & 44.5 \\
BGE-M3 (Default) & 0.118 & \textbf{0.82} & \textbf{45.5} \\
Qwen3-Embedding-0.6B & \textbf{0.116} & \textbf{0.82} & 45.4 \\
E5-Mistral-7B & 0.117 & \textbf{0.82} & \textbf{45.5} \\
\bottomrule
\end{tabular}
\caption{\textbf{Comparison of frozen text encoders} for extracting the $K{=}32$ whitened skill coordinates $\mathbf{z}_i$.}
\label{tab:encoder}
\end{table}

\section{Extended Discussion}
\label{app:discussion}

\paragraph{Why bilinear factorization beats graph clustering and message passing.}
An alternative way to share rollout feedback across prompts is to partition the corpus into discrete clusters or propagate labels along a nearest-neighbor graph. Bilinear skill--difficulty factorization offers three concrete advantages over discrete clustering and graph propagation:
\begin{enumerate}[leftmargin=*,itemsep=1.5pt,topsep=2pt]
\item \emph{Compositional skill representation}: A competition math problem frequently combines two or more skills (e.g., a combinatorics problem that reduces to a modular arithmetic recurrence). In a hard or soft cluster model, a prompt is convexly averaged over cluster centroids, which cannot represent additive skill composition. In \method{}, the inner product $\mathbf{z}_i^\top\boldsymbol{\theta}_t$ naturally sums proficiency contributions across all active coordinates of $\mathbf{z}_i$.
\item \emph{Explicit separation of global drift from item hardness}: Graph label propagation smooths total pass rates across neighbors; if two neighboring problems have the same topic but one has a subtle algebraic trap, graph smoothing pulls their estimates together and blurs within-neighborhood ranking. By keeping $b_i$ outside the subspace projection $\mathbf{z}_i^\top\boldsymbol{\theta}_t$ and updating $b_i$ with long memory ($\gamma_b{=}0.985$), \method{} preserves sharp local hardness contrasts indefinitely.
\item \emph{No iterative inner loop or graph tuning}: Variational message passing on a graph requires multiple coordinate-ascent sweeps per step and tuning neighborhood size and coupling temperature. \method{} performs a single conjugate $32\times 32$ linear solve in $0.54$\,ms.
\end{enumerate}

\paragraph{When does \method{} reduce to simpler estimators?}
If $K{=}0$, \method{} reduces to an independent per-prompt P\'olya--Gamma filter on $b_i$, which is the logistic state-space counterpart of per-item dynamics-predictive tracking in DPS~\citep{mao2026dynamics}. If $\gamma_b{=}0$ and $b_i{\equiv}0$, \method{} reduces to a contextual linear bandit in the embedding space, which captures coarse topic competence similar to zero-shot prompt predictors~\citep{qu2026small} but cannot distinguish easy from hard problems within the same topic (as seen in the $b_i{\equiv}0$ row of Table~\ref{tab:ablation}). Both components are needed to jointly minimize Brier calibration error and maximize Spearman rank correlation, providing a reliable signal for downstream group-prioritized training~\citep{mao2026rlvr} and rollout budget allocation~\citep{zou2026trace}.